\documentclass[10pt,twocolumn,letterpaper]{article}

\pdfoutput=1
\usepackage{wacv}
\usepackage{times}
\usepackage{epsfig}
\usepackage{graphicx}
\usepackage{amsmath}
\usepackage{amssymb}
\usepackage{booktabs}
\usepackage{pifont}
\newcommand{\cmark}{\ding{51}}%
\newcommand{\xmark}{\ding{55}}%
\usepackage{bigdelim}
\usepackage{microtype}
\usepackage{multirow}
\usepackage{multicol}
\usepackage{color, colortbl}
\usepackage{booktabs}
\definecolor{Gray}{gray}{0.90}
\definecolor{Pink}{cmyk}{0, 0.25, 0.1, 0.1}
\def\wacvPaperID{209} 

\wacvfinalcopy 

\ifwacvfinal
\usepackage[breaklinks=true,bookmarks=false]{hyperref}
\else
\usepackage[pagebackref=true,breaklinks=true,colorlinks,bookmarks=false]{hyperref}
\fi

\begin{document}

\title{Motion Aware Self-Supervision for Generic Event Boundary Detection}

\author{Ayush K. Rai, Tarun Krishna, Julia Dietlmeier, Kevin McGuinness$^*$, Alan F. Smeaton$^*$, Noel E. O'Connor$^*$\\
Insight SFI Centre for Data Analytics, Dublin City University (DCU) \\
{\tt\small ayush.rai3@mail.dcu.ie}
}

\maketitle
\def\thefootnote{*}\footnotetext{Equal supervision}\def\thefootnote{\arabic{footnote}}
\thispagestyle{empty}

\begin{abstract}
The task of Generic Event Boundary Detection (GEBD) aims to detect moments in videos that are naturally perceived by humans as generic and taxonomy-free event boundaries. Modeling the dynamically evolving temporal and spatial changes in a video makes GEBD a difficult problem to solve. Existing approaches involve very complex and sophisticated pipelines in terms of architectural design choices, hence creating a need for more straightforward and simplified approaches. In this work, we address this issue by revisiting a simple and effective self-supervised method and augment it with a differentiable motion feature learning module to tackle the spatial and temporal diversities in the GEBD task. We perform extensive experiments on the challenging Kinetics-GEBD and TAPOS datasets to demonstrate the efficacy  of the proposed approach compared to the other self-supervised state-of-the-art methods. We also show that this simple self-supervised approach learns motion features without any explicit motion-specific pretext task. Our results can be reproduced on \href{https://github.com/rayush7/motion_ssl_gebd}{\textcolor{red}{github}}. 
\end{abstract}

\begin{figure*}
    \centering
    \includegraphics[width=1.0\textwidth]{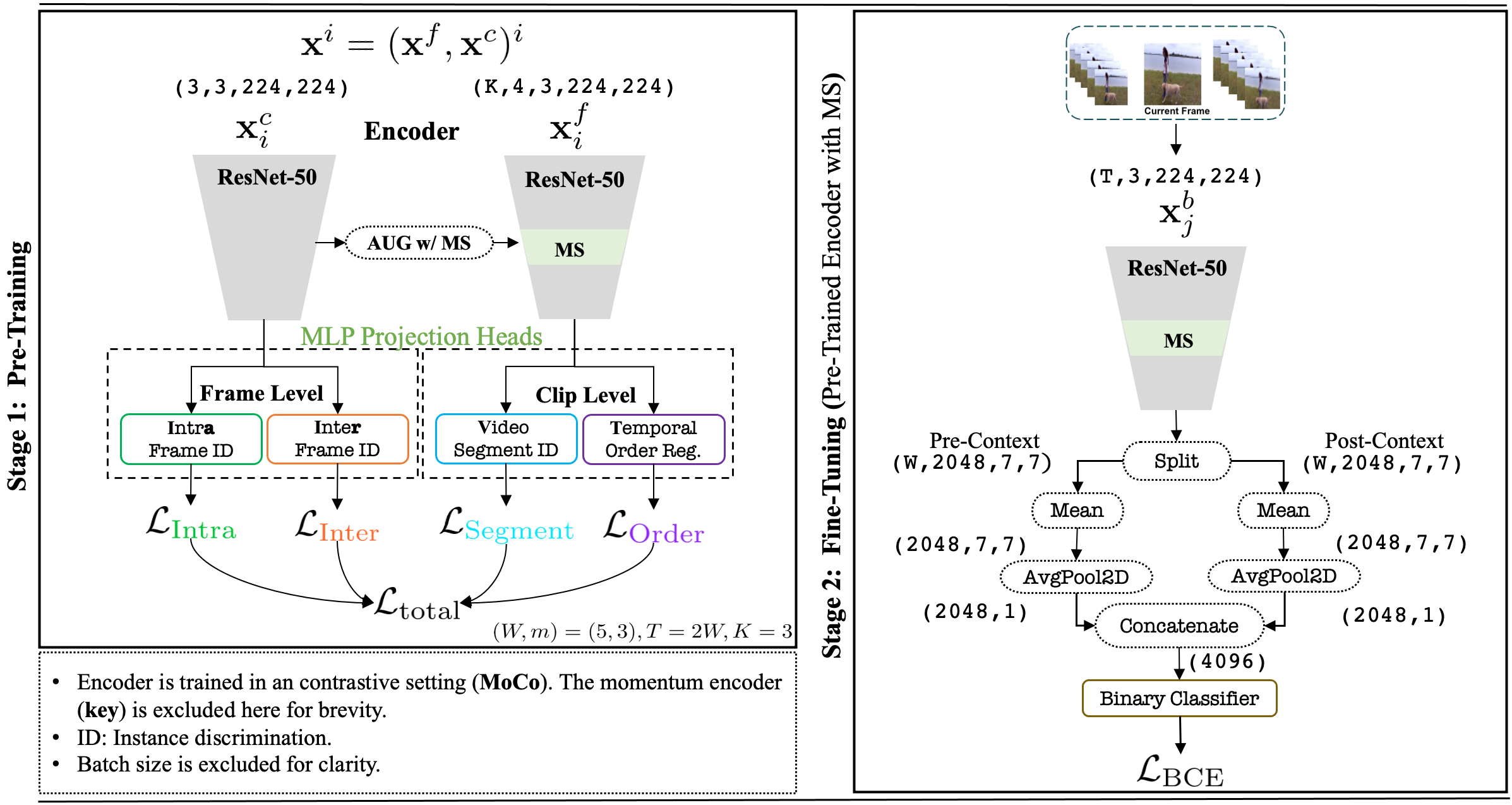}
    \caption{The overall architecture consists of two stages: a) \textbf{Stage 1} involves the pre-training of the modified ResNet50 encoder (augmented with a \textit{MotionSqueeze} layer) with four pretext tasks using a contrastive learning based objective; b) \textbf{Stage 2} consists of fine tuning of the encoder on the downstream GEBD task. Refer to Table \ref{tab:arch_label} in Supplementary material for encoder details.} 
    \label{fig:arch}
\end{figure*}

\section{Introduction}
Modeling videos using deep learning methods in order to learn effective global and local video representations is an extremely challenging task. Current state-of-the-art video models \cite{feichtenhofer2019slowfast} are built upon a limited set of predefined action classes and usually process short clips followed by a pooling operation to generate global video-level predictions. Other mainstream computer vision tasks for video processing have been mainly focused on action anticipation \cite{Miech_2019_CVPR_Workshops,abu2018will}, temporal action detection \cite{Chao_2018_CVPR,gao2017cascaded}, temporal action segmentation \cite{10.1007/978-3-319-46487-9_3,kuehne2014language} and temporal action parsing \cite{Pirsiavash_2014_CVPR,Shao_2020_CVPR}.  However, only limited attention has been given to understanding long form videos. Cognitive scientists \cite{tversky2013event} have observed that humans perceive videos by breaking them down into shorter temporal units, each carrying a semantic meaning and can also reason about them. This creates a need to investigate research problems to detect temporal boundaries in videos that is consistent with their semantic validity and interpretability from a cognitive point of view. 

To this end, the GEBD task was recently introduced in \cite{shou2021generic}\footnote{\label{loveu}\href{https://sites.google.com/view/loveucvpr21/challenge-intro?authuser=0}{LOVEU@CVPR2021}, \href{https://sites.google.com/view/loveucvpr22/program?authuser=0}{LOVEU@CVPR2022}} with an objective to study the long form video understanding problem through the lens of a human perception mechanism. GEBD aims at identifying changes in content, independent of changes in action, brightness, object, etc.,  i.e.~generic event boundaries, making it different to tasks such as video localization \cite{9062498}. Video events could indicate completion of goals or sub-goals, or occasions where it becomes difficult for humans to predict what will happen next. The recently released Kinetics-GEBD dataset \cite{shou2021generic} is the first dataset specific to the GEBD task. It is annotated by 5 different event boundary annotators, thereby capturing the subtlety involved in human perception and making it the dataset with the greatest number of temporal boundaries ($8 \times$  EPIC-Kitchen-100 \cite{damen2018scaling} and $32 \times$  ActivityNet \cite{caba2015activitynet}). The primary challenge in the GEBD task is to effectively model generic spatial and temporal diversity as described in DDM-Net \cite{tang2021progressive}. Spatial diversity is primarily the result of both low-level changes, e.g.\ changes in brightness or appearance, and high-level changes, e.g., changes in camera angle, or appearance and disappearance of the dominant subject. Temporal diversity, on the other hand, can be attributed to changes in action or changes by the object of interaction with different speeds and duration, depending on the subject. These spatio-temporal diversities make GEBD a difficult problem to address. 

In this work, to address the biased nature of video models trained over predefined classes in a supervised setting, and the spatial diversity in GEBD, we leverage the power of self-supervised models. Self-supervised techniques like TCLR \cite{dave2022tclr} and CCL \cite{kong2020cycle} have achieved breakthrough results on various downstream tasks for video understanding. The representations learned using self-supervised learning (SSL) methods are not biased towards any predefined action class making SSL methods an ideal candidate for the GEBD task.  In addition, in order to characterize temporal diversity in GEBD, learning motion information is essential to capture the fine-grained temporal variations that occur during the change of action scenarios. Previous methods in video modeling learn temporal motion cues by pre-computing the optical flow \cite{lin2018bsn,lin2019bmn,liu2018future} between consecutive frames, which is done externally and requires substantial computation. Alternatively, methods such as those described in \cite{ilg2017flownet,fischer2015flownet} estimate optical flow internally by learning visual correspondences between images. The motion features learnt on-the-fly can also be used for downstream applications such as action recognition as  illustrated in \cite{zhao_recog_cvpr18, kwon2020motionsqueeze}. 

This presents an interesting research question: how can we develop a SSL framework for video understanding that accounts for both appearance and motion features? Do we need an explicit motion-specific training objective or can this be implicitly achieved? We answer these questions by rethinking SSL by reformulating the training objective proposed in VCLR~\cite{kuang2021video} at clip-level and further integrating it with a differentiable motion estimation layers using the \textit{MotionSqueeze} (MS) module introduced in \cite{kwon2020motionsqueeze} to jointly learn appearance and motion features for videos. To summarise, the main contributions of our work are as follows:

\begin{itemize}
 
    \item We revisit a simple self-supervised method VCLR \cite{kuang2021video} with a noticeable change by modifying its pretext tasks by splitting them into frame-level and clip-level to learn effective video representations (cVCLR) . We further augment the encoder with a differentiable motion feature learning module for GEBD.
    \item We conduct exhaustive evaluation on the Kinetics-GEBD and TAPOS datasets and show that our approach achieves comparable performance to the self-supervised state-of-the-art methods without using enhancements like model ensembles, pseudo-labeling or the need for other modality features (e.g.~audio).
    \item We show that the model can learn motion features under self-supervision even without having any explicit motion specific pretext task. 
\end{itemize}


\section{Related Work}
\subsection{Generic Event Boundary Detection. }
The task of GEBD \cite{shou2021generic} is similar in nature to the Temporal Action Localization (TAL) task, where the goal is to localize the start and end points of an action occurrence along with the action category. Initial attempts to address GEBD were inspired from popular TAL solvers including the boundary matching networks (BMN) \cite{lin2019bmn} and BMN-StartEnd \cite{shou2021generic}, which generates proposals with precise temporal boundaries along with reliable confidence scores. Shou \textit{et al.}~\cite{shou2021generic} introduced a supervised baseline Pairwise Classifier (PC),  which considers GEBD as a framewise binary classification problem (boundary or not) by having a simple linear classifier that uses concatenated average features around the neighbourhood of a candidate frame. However, since GEBD is a new task, most of the current methods are an extension of state-of-the-art video understanding tasks, which overlook the subtle differentiating characteristics of GEBD. Hence there is a necessity for GEBD specialized solutions. 

DDM-Net \cite{tang2021progressive} applied progressive attention on multi-level dense difference maps (DDM) to characterize motion patterns and jointly learn motion with appearance cues in a supervised setting. However, we learn generic motion features by augmenting the encoder with a MS module in a self-supervised setting. Hong \textit{et al.}~\cite{hong2021generic} used a cascaded temporal attention network for GEBD, while Rai \textit{et al.}~\cite{rai2021discerning} explored the use of spatio-temporal features using two-stream networks. Li \textit{et al.}~\cite{Li_2022_CVPR} designed an end-to-end spatial-channel compressed encoder and temporal contrastive module to determine event boundaries. Recently, SC-Transformer \cite{li2022structured} introduced a structured partition of sequences (SPoS) mechanism to learn structured context using a transformer based architecture for GEBD and augmented it with the computation of group similarity to learn distinctive features for boundary detection. One advantage of SC-Transformer is that it is independent of video length and predicts all boundaries in a single forward pass by feeding in 100 frames, however it requires substantial memory and computational resources.

Regarding unsupervised GEBD approaches, a shot detector library\footnote{ \url{https://github.com/Breakthrough/PySceneDetect}} and PredictAbility (PA) have been investigated in \cite{shou2021generic}. The authors of UBoCo \cite{kang2021uboco,kang2021winning} proposed a novel supervised/unsupervised method that applies contrastive learning to a TSM\footnote{\label{tsmatrix} Temporal Self-Similarity Matrix} based intermediary representation of videos to learn discriminatory boundary features. UBoCo's recursive TSM\textsuperscript{\ref{tsmatrix}} parsing algorithm exploits generic patterns and detects very precise boundaries. However, they pre-process all the videos in the dataset to have the same frames per second (fps) value of 24, which adds a computational overhead. Furthermore, like the SC-Transformer, UBoCo inputs the frames representing the whole video at once, whereas in our work we use raw video signals for pre-training and only the context around the candidate boundary as input to the GEBD task. TeG \cite{qian2021exploring} proposed a generic self-supervised model for video understanding for learning persistent and more fine-grained features and evaluated it on the GEBD task. The main difference between TeG and our work is that TeG uses a 3D-ResNet-50 encoder as their backbone, which makes the training computationally expensive, whereas we use 2D-ResNet-50 model and modify it by adding temporal shift module (TSM\footnote{\label{temp_shift}Temporal Shift Module}) \cite{lin2019tsm} to achieve the same effect as 3D convolution while keeping the complexity of a 2D CNN.



GEBD can be used as a preliminary step in a larger downstream application, e.g.~video summarization, video captioning \cite{wang2022generic}, or ad cue-point detection \cite{chen2021shot}. It is, therefore, important that the GEBD model not add excessive computational overhead to the overall pipeline, unlike many of the examples of related work presented here.

\subsection{SSL for video representation learning. }
Self-supervision has become the new norm for learning representations given its ability to exploit unlabelled data \cite{noroozi2016unsupervised,gidaris2018unsupervised, doersch2015unsupervised, asano2019self, caron2020unsupervised, zbontar2021barlow,bardes2021vicreg, chen2020simple,oord2018representation, krishna2021evaluating, Djilali_2021_ICCV}.
Recent approaches devised for video understanding can be divided into two categories based on the SSL objective, namely pretext task based and contrastive learning based.

\textbf{Pretext task based.} The key idea here is to design a pretext task for which labels are generated in an online fashion, referred to as \textit{pseudo labels}, without any human annotation. Examples include: predicting correct temporal order \cite{misra2016shuffle}, Video Rot-Net \cite{jing2018self} for video rotation prediction, clip order prediction \cite{xu2019self}, odd-one-out networks \cite{fernando2017self}, sorting sequences \cite{lee2017unsupervised}, and pace prediction \cite{wang2020self}\footnote{leverages contrastive learning as an additional objective as well.}. All these approaches exploit raw spatio-temporal signals from videos in different ways based on pretext tasks and consequently learn representations suitable for varied downstream tasks. 


\textbf{Contrastive learning based.} Contrastive learning approaches bring semantically similar objects, clips, etc., close together in the embedding space while contrasting them with negative samples, using objectives based on some variant of Noise Contrastive Estimation (NCE) \cite{pmlr-v9-gutmann10a}.  The Contrastive Predictive Coding (CPC) approach \cite{oord2018representation} for images was extended to videos in DPC \cite{han2019video} and MemDPC \cite{han2020memory}, which augments DPC with the notion of \textit{compressed memory}. Li \textit{et al.}~\cite{tao2020self} extends the contrasting mutli-view framework for inter-intra style video  representation, while Kong \textit{et al.}~\cite{kong2020cycle} combine ideas from \textit{cycle-consistency} with contrastive learning to propose \textit{cycle-contrast}. Likewise,  Yang \textit{et al.}~\cite{yang2020video} exploits visual tempo in a contrastive framework to learn spatio-temporal features. Similarly, \cite{dave2022tclr, bai2020can} use temporal cues with contrastive learning.  VCLR \cite{kuang2021video} formulates a video-level contrastive objective to capture global context. In the work presented here, we exploit VCLR as our backbone objective. However, different to pretext tasks in VCLR, which perform computation only on frame level, we modify those pretext tasks to not only operate on frame-level but also on clip-level thereby leading to better modeling of the spatio-temporal features in videos. See \cite{ChantrySchiappa2022SelfSupervisedLF} for  a more extensive review of SSL methods for video understanding. 



\subsection{Motion estimation and Learning visual correspondences for video understanding. }
\textbf{Motion estimation. } Two-stream architectures \cite{feichtenhofer2016convolutional,simonyan2014two} have exhibited promising performance on the action recognition task by using pre-computed optical flow, although such approaches reduce the efficiency of video processing. Several other methods \cite{fan2018end,jiang2019stm,lee2018motion,piergiovanni2019representation,sun2018optical} have proposed architectures that learn motion internally in an end-to-end fashion. The work presented in \cite{li2021motion,yang2020hierarchical} introduced a motion-specific contrastive learning task to learn motion features in a self-supervised setting.

\textbf{Learning visual correspondences. }
Many recent works have  proposed to learn visual correspondences between images using neural networks \cite{fischer2015flownet,han2017scnet,lee2019sfnet,min2019hyperpixel,rocco2017convolutional,sun2018pwc}. Regarding learning correspondences for video understanding, CPNet \cite{liu2019learning} introduced a network that learns representations of videos by mixing appearance and long-range motion features from an RGB input only. Zhao \textit{et al.}~\cite{zhao_recog_cvpr18} proposed a method that learns a disentangled representation of a video, namely static appearance, apparent motion and appearance change from  RGB input only. \textit{MotionSqueeze} (MS) \cite{kwon2020motionsqueeze} introduced an end-to-end trainable, model-agnostic and lightweight module to extract motion features that does not require any correspondence supervision for learning. 


\section{Method}
In order to apply a contrastive learning framework to videos specifically for generic event boundary detection, we follow the framework proposed by VCLR \cite{kuang2021video} and make noticeable modifications to it. For simplicity the notations are kept similar to \cite{kuang2021video} unless otherwise explicitly stated.


\subsection{SSL for Video Representation Learning}
\noindent \textbf{1: Contrastive encoder.} Our processing backbone is a ResNet-50 based encoder equipped with four pretext tasks, as defined in VCLR \cite{kuang2021video}, trained following a contrastive objective as defined in MoCo-V2 \cite{chen2020improved}.  

Let $\mathbf{x}_{p} = \mathcal{T}(\mathbf{x}_{q}$) be an augmented view of an anchor image 
$\mathbf{x}_{q}$ with $\mathcal{T} \sim \mathcal{P}$ ($\mathcal{P} =$ \{random scaling, color-jitter, random grayscale, random Gaussian blur, and random horizontal flip\} being set of augmentations) and $\mathcal{N}^{-}$ negative samples.  $\mathbf{x}_{q}$ and $\mathbf{x}_{p}$
are processed through query ($f_{q}(\mathbf{x}_{q})$) and a key $(f_{k}(\mathbf{x}_{p}))$ 
encoder respectively. In addition, these encoders are  appended with projection heads (MLP layers) to get low dimensional representations of inputs i.e. $q= g_{q}(f_{q}(\mathbf{x}_{q})),  p =g_{k}(f_{k}(\mathbf{x}_{p}))$. The overall objective can be optimized by InfoNCE loss \cite{oord2018representation}:
%
\begin{equation}
   \mathcal{L}_\text{NCE}(q,p,\mathcal{N^-}) =  -\log\Bigg( \frac{ e^{ \text{sim} ({q,p})} } {e^{ \text{sim} ({q,p})} + \sum_{j=1}^{\text{N}} e^{ \text{sim} ({q,n_j})} }\Bigg),
\end{equation}
where, $\mathrm{sim}(\cdot, \cdot)$ is a similarity function. We note that $g_{q}$ and $g_{k}$ can be thought of as a task-specific projection heads with further  details below\footnote{subscript $q$, $k$ in $f$ and $g$ represents \textit{query} and \textit{key}}. 

\noindent \textbf{2: Pre-text setup.}
In order to capture different generic subtle nuances (spatial variations, temporal coherency, long range  dependencies) for video understanding, the pretext tasks defined in VCLR \cite{kuang2021video} are a good candidate for pre-training as they serve the purpose of capturing such semantics for powerful video representation from raw video signals. We alter the pretext task setup in VCLR to ensure that Intra and Inter instance discrimination (ID) tasks operate at frame-level while computation of the video segment ID and temporal order regularization tasks occurs at clip-level. Below we elaborate on the intuition behind this notion. 


For frame-level pretext tasks, consider three randomly selected frames from a video, $v_1$, $v_2$ and $v_3$. $v_1$ undergoes different augmentations to generate $v_1^a$ and $v_1^+$. $\mathcal{N}^-$ represents negative samples from other videos. $v_1^a$ is processed through query encoder $f_{q}(v_1^a)$, while ($v_1^+$, $v_2$, $v_3$) is processed through key encoder $f_{k}(\cdot)$. While depending upon the pretext, the projection head varies across the tasks. 

For clip-level pretext tasks, a video $V$ is divided into $K$ (set to 3) segments $\{S_1,S_2,...,S_K\}$ of equal duration. Two tuples, comprising of 4 frame long clip, are randomly and independently sampled from each of these segments to form an anchor tuple and a positive tuple. For instance, let $c_k = \{u_1,u_2,u_3,u_4\}$ where $c_k$ denote the ordered clip sampled from the $k^\text{th}$ segment while $u_1.....u_4$ represent the frames in that clip. Similarly the anchor and positive tuple are given by $t^a=\{c_1^a,c_2^a, \ldots c_K^a\}$ and $t^+=\{c_1^+,c_2^+, \dots c_K^+\}$ respectively.

\textbf{a. Intra-frame ID task.}
In order to model spatial diversity for the GEBD task, we adopt the intra-frame instance discrimination task proposed in VCLR \cite{kuang2021video} to model inherent spatial changes across frames. 
 For this task only $v_1^+$ is considered as a positive example while $v_2$ and $v_3$ represent negative examples. MLP heads are given by $g_q^r$ and $g_k^r$ while anchor embedding $q_1^a = g_q^r(f_q(v_1^a))$, positive embedding as $p_1^+ = g_k^r(f_k(v_1^+))$ and the negative sample\footnote{\textit{Note}: Negative samples comes from the same video i.e.~two samples as shown in Eq.~\eqref{eq3}.} embeddings $p_2 = g_k^r(f_k(v_2))$, $p_3 = g_k^r(f_k(v_3))$. The loss objective is given by:
\begin{equation}\label{eq3}
   \mathcal{L}_\text{Intra} = \mathcal{L}_\text{NCE}(q_1^a,p_1^+,\{p_2,p_3\}).
\end{equation}

\textbf{b. Inter-frame ID task.}
Detecting generic event boundaries requires encoding fine-grained temporal structures from a coherent action, which are consistent with each other.
To model this, inter-frame instance discrimination task considers $v_1^a$ as an anchor frame and ($v_1^+$, $v_2$, $v_3$) as positive samples while $\mathcal{N}^-$ as negative samples. $g_q^e$ and $g_k^e$ are MLP projection heads which output the anchor embedding $q_1^a = g_q^e(f_q(v_1^a))$ and positive embeddings as $p_1^+ = g_k^e(f_k(v_1^+))$, $p_2 = g_k^e(f_k(v_2))$, $p_3 = g_k^e(f_k(v_3))$. Let $p^{\prime} \in \{ p_1^+,p_2,p_3\}$, hence the inter objective becomes:
\begin{align}
\mathcal{L}_\text{Inter} = \frac{1}{3} \sum_{p^\prime}^{}\mathcal{L}_\text{NCE}(q_1^a,p^{\prime},\mathcal{N^-}).
\end{align}


\textbf{c. Video segment based ID task.}
Learning long range temporal diversity in a video is also crucial for the GEBD task. For capturing the evolving semantics in the temporal dimension we need to incorporate global video level information. 
The contrastive loss objective is chosen in a way that each clip in the clip anchor and clip positive tuples i.e. $t^a$ and $t^+$ learn a video-level embedding through consensus operation (denoted by $\mathcal{C}$) e.g.~average. Mathematically this can be represented as:
\begin{equation}
    q_t^a = g_p^s(\mathcal{C}\mid \mathcal{C}(f_q(c_1^a)),\mathcal{C}(f_q(c_2^a)), \ldots ,\mathcal{C}(f_q(c_K^a))),
\end{equation}
\begin{equation}
    p_t^+ = g_k^s(\mathcal{C}\mid \mathcal{C}(f_k(c_1^+)),\mathcal{C}(f_k(c_2^+)), \ldots ,\mathcal{C}(f_k(c_K^+))),
\end{equation}
\begin{equation}
   \mathcal{L}_\text{Segment} = \mathcal{L}_\text{NCE}(q_t^a,p_t^+,\mathcal{N}^-).
\end{equation}
\noindent 
Here, $g_q^s$ and $g_k^s$ represent the MLP heads, $\mathcal{C}(f_q(c_k^a))$ indicates the average over the encoder representation of the individual frame in the $k^\text{th}$ clip while $q_t^a$, $p_t^+$ denotes the final embeddings for anchor and positive clip tuples. The video-level contrastive loss is given by $\mathcal{L}_\text{Segment}$.

\textbf{d. Temporal order regularization task.}
In order to enforce inherent sequential structure on videos for signalling supervision in self-supervised video representation learning, we need a pretext task to learn the correct temporal order of the video data. This can also be attained through pretext tasks proposed in \cite{fernando2017self,wang2020self}. However, in this work we restrict ourselves to use the temporal ordering as a regularization term (denoted by $\mathcal{L}_\text{Order}$) within the contrastive framework as explained in Section 3.3 in \cite{kuang2021video} though we reformulate it to include clip-level computation.

\noindent The modifications made to video segment based ID task and temporal order regularization task in VCLR to incorporate clip-level computation is referred to as \texttt{cVCLR} (clip-VCLR).

\subsection{Motion Estimation}
\label{sec:motionestimation}

For learning motion features we use the \textit{MotionSqueeze} (MS) module presented in \cite{kwon2020motionsqueeze},  a learnable motion feature extractor that can be inserted into any video understanding architecture to learn motion features and replace the external computation of optical flow. The motion features are learned in three steps:

\noindent \textbf{1: Correlation computation.}
Consider $F^{(t)}$ and $F^{(t+1)}$ represent two adjacent input feature maps of spatial resolution $H\times W$ and channel dimension $C$. 
The correlation tensor $\mathtt{S}^{(t)}$ is computed by calculating correlation score of every spatial position $\mathtt{x}$ with respect to displacement $\mathtt{p}$ following the correlation layer implementation in FlowNet \cite{fischer2015flownet}. The correlation for position $\mathtt{x}$ is only computed in neighborhood size $\mathtt{P}=2l+1$ by restricting a maximum displacement $\mathtt{p} \in [-l,l]^2$ and the value of $\mathtt{P}$ is set to $15$.


\noindent \textbf{2: Displacement estimation.}
The next step involves estimating the displacement map of size $H\times W\times2$ from the correlation tensor $\mathtt{S}^{(t)}$. To get the best matching displacement for position $\mathtt{x}$, \textit{kernal-soft-argmax} \cite{lee2019sfnet} is used. 
In addition, a motion confidence map (of size $H\times W\times1$) of correlation as auxiliary motion information is obtained by pooling the highest correlation on each position $\mathtt{x}$ as described in \cite{kwon2020motionsqueeze}.
The motion confidence map helps in identifying displacement outliers and learn informative motion features. The displacement map is then concatenated with the motion confidence map to create a displacement tensor $\mathbf{D}^{(t)}$  of size $H \times W \times 3$.

\noindent \textbf{3: Feature transformation. }
In order to convert displacement tensor $\mathbf{D}^{(t)}$ to a relevant motion feature $\mathbf{M}^{(t)}$ (with same channel dimension $C$ as input $\mathbf{F}^{(t)}$) , $\mathbf{D}^{(t)}$ is passed through four depth-wise separable convolutions \cite{howard2017mobilenets} similar to \cite{kwon2020motionsqueeze}. 
Contrary to \cite{kwon2020motionsqueeze}, in our work, we apply this feature transformation in a self-supervision setting to learn displacement tensor and motion confidence map (generic motion features). Finally, the motion features $\mathbf{M}^{(t)}$ are added to the input of the next layer using an element-wise addition operation : $\mathbf{F}^{\prime(t)} = \mathbf{F}^{(t)} + \mathbf{M}^{(t)}$. The resulting fused feature $\mathbf{F}^{\prime(t)}$ is passed as  input to the next layer. 
For more details we refer the readers to \cite{kwon2020motionsqueeze}.



\subsection{\label{optim}Optimisation}
The  overall contrastive \textit{loss objective} is given by:
\begin{equation}\label{eq9}
    \mathcal{L}_\text{total} = \mathcal{L}_\text{Inter} + \mathcal{L}_\text{Intra} + \mathcal{L}_\text{Segment} + \mathcal{L}_\text{Order}.
\end{equation}
\noindent 
Our encoder is augmented with a MS module (introduced after $\texttt{conv3\_x}$\footnote{notation as in \cite{he2016deep}}) to jointly learn appearance and motion features. More precisely, a ResNet-50 \cite{he2016deep} model is adopted as the CNN encoder and we insert a TSM\textsuperscript{\ref{temp_shift}} \cite{lin2019tsm} for each residual block of ResNet. 
Each of the four losses contribute equally to $\mathcal{L}_\text{total}$ although weighing them appropriately might boost the performance. The overall framework is illustrated in Figure~\ref{fig:arch}.

\begin{table*}[t]
    
    \centering
    \caption{F1 scores on the Kinetics-GEBD validation set with Rel. Dis threshold ranging from 0.05 to 0.5 with step of 0.05. $\ddagger$: soft-labels, $\dagger$: hard-labels. $^\mathbf{\ast}$ is pretrained on Kinetics-400~\cite{kay2017kinetics} dataset.}
    \label{tab:results_kinetics-gebd}
    \resizebox{17cm}{!}{
    \begin{tabular}{llccccccccccc|c}
    \toprule
        \multicolumn{2}{c}{Rel.\ Dis Threshold} & Finetuning & 0.05 & 0.1 & 0.15 & 0.2 & 0.25 & 0.3 & 0.35 & 0.4 & 0.45 & 0.5 & avg \\
    \midrule
    & BMN$^{\dagger}$ \cite{lin2019bmn} & \xmark&0.186 & 0.204 & 0.213 & 0.220 & 0.226 & 0.230 & 0.233 & 0.237 & 0.239 & 0.241 & 0.223 \\
    & BMN-SE \cite{lin2019bmn}$^{\dagger}$ &\xmark &0.491 & 0.589 & 0.627 & 0.648 & 0.660 & 0.668 & 0.674 & 0.678 & 0.681 & 0.683 & 0.640 \\
    & TCN-TAPOS \cite{lea2016segmental}$^{\dagger}$ & \cmark & 0.464 & 0.560 & 0.602 & 0.628 & 0.645 & 0.659 & 0.669 & 0.676 & 0.682 & 0.687 & 0.627 \\
    & TCN \cite{lea2016segmental}$^{\dagger}$ & \xmark & 0.588 & 0.657 & 0.679 & 0.691 & 0.698 & 0.703 & 0.706 & 0.708 & 0.710 & 0.712 & 0.685\\
    Supervised & PC \cite{shou2021generic}$^{\dagger}$  & \cmark&0.625 & 0.758 & 0.804 & 0.829 & 0.844 & 0.853 & 0.859 & 0.864 & 0.867 & 0.870 & 0.817 \\
    & SBoCo-Res50 \cite{kang2021uboco}$^{\dagger}$  & \xmark & 0.732 & - & - & - & - & - & - & - & - & - & 0.866 \\
    & SBoCo-TSN \cite{kang2021uboco}$^{\dagger, \mathbf{\ast}}$  & \xmark & 0.787 & - & - & - & - & - & - & - & - & - & 0.892 \\
    & DDM-Net \cite{tang2021progressive}$^{\dagger}$ & \xmark&0.764 & 0.843 & 0.866 & 0.880 & 0.887 & 0.892 & 0.895 & 0.898 & 0.900 & 0.902 & 0.873\\
    & Li et.al.~\cite{Li_2022_CVPR}$^{\ddagger}$ & \xmark&0.743 & 0.830 & 0.857 & 0.872 & 0.880 & 0.886 & 0.890 & 0.893 & 0.896 & 0.898 & 0.865 \\
    & \textbf{SC-Transformer} \cite{li2022structured}$^{\ddagger}$ & \xmark&\textbf{0.777} & \textbf{0.849} & \textbf{0.873} & \textbf{0.886} & \textbf{0.895} & \textbf{0.900} & \textbf{0.904} & \textbf{0.907} & \textbf{0.909} & \textbf{0.911} & \textbf{0.881}\\
    \midrule
    & SceneDetect \cite{pyscene_det} & \xmark & 0.275 & 0.300 & 0.312 & 0.319 & 0.324 & 0.327 & 0.330 & 0.332 & 0.334 & 0.335 & 0.318 \\
    & PA - Random \cite{shou2021generic} & \xmark& 0.336 & 0.435 & 0.484 & 0.512 & 0.529 & 0.541 & 0.548 & 0.554 & 0.558 & 0.561 & 0.506 \\
    & PA \cite{shou2021generic} & \xmark&0.396 & 0.488 & 0.520 & 0.534 & 0.544 & 0.550 & 0.555 & 0.558 & 0.561 & 0.564 & 0.527 \\
     Un-supervised & UBoCo-Res50 \cite{kang2021uboco} & \xmark & 0.703 & - & - & - & - & - & - & - & - & - & 0.866 \\
    & UBoCo-TSN \cite{kang2021uboco}$^{\mathbf{\ast}}$ & \xmark & 0.702 & - & - & - & - & - & - & - & - & - & 0.892 \\
    \rowcolor{Gray}
   & TeG-PS \cite{qian2021exploring}$^{\dagger}$  & \cmark &0.699 & - & - & - & - & - & - & - & - & - & - \\
    \rowcolor{Gray}
   (Self-supervised)& TeG-FG \cite{qian2021exploring}$^{\dagger}$  & \cmark &0.714 & - & - & - & - & - & - & - & - & - & - \\
    \rowcolor{Gray}
    & Ours$^{\dagger}$  & \cmark& 0.680 & 0.779 & 0.806 & 0.818 & 0.825 & 0.830 & 0.834 & 0.837 & 0.839 & 0.841 & 0.809 \\
    \rowcolor{Gray}
    & Ours$^{\ddagger}$  & \cmark&0.711 & 0.777 & 0.791 & 0.795 & 0.798 & 0.799 & 0.801 & 0.802 & 0.802 & 0.803 & 0.788 \\
    
    \bottomrule
    \end{tabular}
    }
\end{table*}

\section{Experimental Setup}

\subsection{Implementation Details.}
\noindent \textbf{Stage 1: Pre-training.} We closely follow VCLR \cite{kuang2021video} to train  the encoder.
The model was pre-trained end-to-end with the objective as defined in Eq.~\eqref{eq9} on 2 NVIDIA GeForce RTX-2080Ti GPUs with an effective batch size ($\mathcal{B}$) of 8 distributed across the GPUs (4 each) with temperature set to 0.01 across all pretext tasks. The input to the frame level and clip level pretext task is   ($\mathcal{B}, 3, 3, 224, 224$) and ($\mathcal{B}, K, 4, 3, 224, 224$) respectively with $K=3$. TSM\textsuperscript{\ref{temp_shift}} and \textit{MotionSqueeze} is only applied on clip level task. The encoder is initialised to MoCo-v2 \cite{chen2020improved} weights with negative samples $\mathcal{N}^{-}$ (queue size) set to 8192 and is trained with SGD for 400 epochs with a warm-start
of 5 epochs following a cosine decay with base learning rate of 0.01. \textit{Pre-training} is only performed on the Kinetics-GEBD \cite{shou2021generic} dataset.

\noindent \textbf{Stage 2: Finetuning.} Input to the encoder is based on the temporal window ($W$=5) which defines a context over a candidate frame (before and after) with a stride $m=3$\footnote{selecting one frame
out of every 3 consecutive frames} resulting into a 4D tensor $(10,3,224,224)$ as input.
We fine-tune the model end-to-end with a binary cross entropy (BCE) (boundary is 0/1)  as the objective augmented with Gaussian smoothing ($\sigma=3$) for soft labeling as in \cite{li2022structured}. 
The learning rate set to 7.5$e^{-4}$ for Kinetics-GEBD, while for TAPOS it was set to 1$e^{-4}$. 
Balance sampling is applied to each batch during training to avoid class imbalance. We finetune the model for 8 epochs and use early stopping to find the best model. More details in supplementary material.

To select the final boundary predictions for the video,
we apply post-processing scheme on the obtained boundary proposals. \textit{First}, proposals should be greater then a threshold of 0.5. \textit{Second}, we aggregate all the proposals within a 1 second time window.

\begin{table*}[t]
    
    \centering
    \caption{F1 scores on the TAPOS validation set with Rel.\ Dis threshold ranging from 0.05 to 0.5 with step of 0.05. $\ddagger$: soft-labels, $\dagger$: hard-labels. (-) : Not clear.}
    \vspace{5pt}
    \label{tab:results_tapos}
    \resizebox{17cm}{!}{
    \begin{tabular}{llccccccccccc|c}
    \toprule
    \multicolumn{2}{c}{Rel.\ Dis Threshold} & Finetuning & 0.05 & 0.1 & 0.15 & 0.2 & 0.25 & 0.3 & 0.35 & 0.4 & 0.45 & 0.5 & avg \\
    \midrule
    & ISBA \cite{ding2018weakly} &- & 0.106 & 0.170 & 0.227 & 0.265 & 0.298 & 0.326 & 0.348 & 0.369 & 0.376 & 0.384 & 0.314 \\
    & TCN \cite{lea2016segmental} & \xmark& 0.237 & 0.312 & 0.331 & 0.339 & 0.342 & 0.344 & 0.347 & 0.348 & 0.348 & 0.348 & 0.330 \\
    & CTM \cite{huang2016connectionist} & -& 0.244 & 0.312 & 0.336 & 0.351 & 0.361 & 0.369 & 0.374 & 0.381 & 0.383 & 0.385 & 0.350\\
    Supervised & Transparser \cite{Shao_2020_CVPR} &- & 0.289 & 0.381 & 0.435 & 0.475 & 0.500 & 0.514 & 0.527 & 0.534 & 0.540 & 0.545 & 0.474\\
    & PC \cite{shou2021generic}$^{\dagger}$ & \cmark & 0.522 & 0.595 & 0.628 & 0.647 & 0.660 & 0.666 & 0.672 & 0.676 & 0.680 & 0.684 & 0.643 \\
    & DDM-Net \cite{tang2021progressive}$^{\dagger}$ & \xmark & 0.604 & 0.681 & 0.715 & 0.735 & 0.747 & 0.753 & 0.757 & 0.760 & 0.763 & 0.767 & 0.728 \\
    & \textbf{SC-Transformer} \cite{li2022structured}$^{\ddagger}$ & \xmark &\textbf{0.618} & \textbf{0.694} & \textbf{0.728} & \textbf{0.749} & \textbf{0.761} & \textbf{0.767} & \textbf{0.771} & \textbf{0.774} & \textbf{0.777} & \textbf{0.780} & \textbf{0.742} \\
    \midrule
    & SceneDetect \cite{pyscene_det} & \xmark & 0.035 & 0.045 & 0.047 & 0.051 & 0.053 & 0.054 & 0.055 & 0.056 & 0.057 & 0.058 & 0.051\\
    Un-supervised & PA - Random \cite{shou2021generic} & \xmark & 0.158 & 0.233 & 0.273 & 0.310 & 0.331 & 0.347 & 0.357 & 0.369 & 0.376 & 0.384 & 0.314 \\ 
    & PA \cite{shou2021generic} & \xmark & 0.360 & 0.459 & 0.507 & 0.543 & 0.567 & 0.579 & 0.592 & 0.601 & 0.609 & 0.615 & 0.543 \\
    \rowcolor{Gray}
    (Self-supervised) & Ours$^{\dagger}$ & \cmark & 0.573 & 0.614 & 0.639 & 0.656 & 0.669 & 0.679 & 0.687 & 0.693 &  0.700 & 0.704 & 0.661 \\
    \rowcolor{Gray}
     & Ours$^{\ddagger}$ & \cmark & 0.586 & 0.624 & 0.648 & 0.663 & 0.675 & 0.685 & 0.692 & 0.697 &  0.704 & 0.708 & 0.668 \\
    \bottomrule
    \end{tabular}
    }
\end{table*}

\begin{figure*}[t]
    \centering
    \includegraphics[scale=0.35]{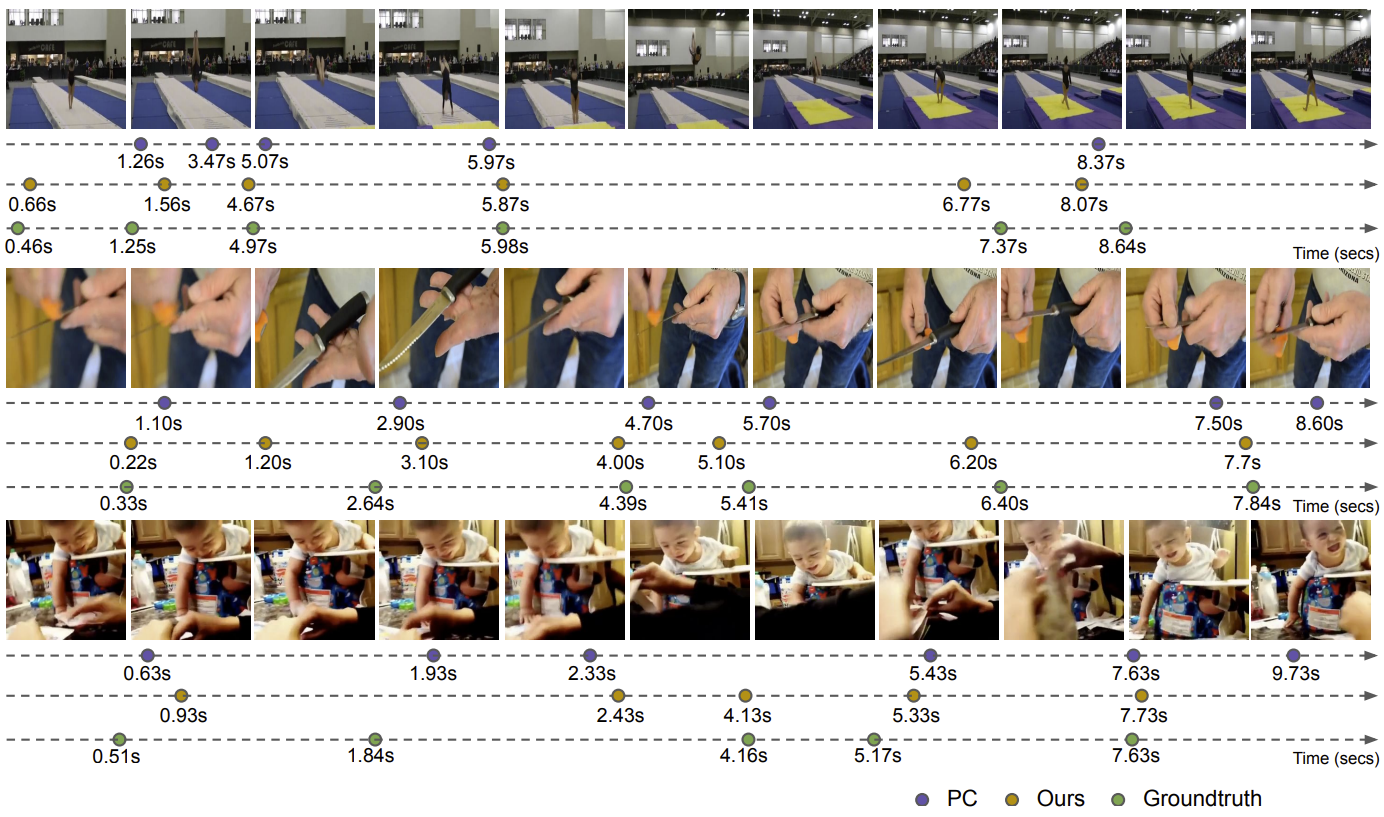}
    \caption{Qualitative Analysis I: visualization of some detected boundaries on the validation set of Kinetics-GEBD. Compared with baseline PC \cite{shou2021generic}, our method produces more precise boundaries that are consistent with the ground truth.}
    \label{fig:bdy}
\end{figure*}




\subsection{Results}
We perform extensive quantitative and qualitative studies on the given datasets. For evaluation protocol we refer the reader to supp. material. In Tables~\ref{tab:results_kinetics-gebd} and \ref{tab:results_tapos}, we report F1 scores for different thresholds ranging from 0.05 to 0.5 with a step of 0.05, for the Kinetics-GEBD and TAPOS datasets respectively. On the Kinetics-GEBD dataset, our model outperforms the supervised baseline PC \cite{shou2021generic} with \textit{Rel.\ Dis} threshold 0.05 and is also comparable with other state-of-the-art unsupervised/self-supervised GEBD models like UBoCo \cite{kang2021uboco} and TeG \cite{qian2021exploring} in terms of performance. Table~\ref{tab:results_kinetics-gebd} illustrates the result on the Kinetics-GEBD dataset. On the TAPOS dataset, which consists of Olympic sport videos with 21 action classes, we have a similar observation. Our model outperforms the supervised baseline PC \cite{shou2021generic} at the \textit{Rel.\ Dis} threshold of 0.05 and is comparable on other thresholds. Other state-of-the-art methods on TAPOS like DDM-Net \cite{tang2021progressive} and SC-Transformer \cite{li2022structured} fall in the supervised category and cannot be directly compared with our results. We were unable to find other state-of-the-art un/self-supervised models for GEBD to directly compare our results with, on the TAPOS dataset shown in Table~\ref{tab:results_tapos}. 

\begin{figure*}
    \centering
    \includegraphics[scale=0.35]{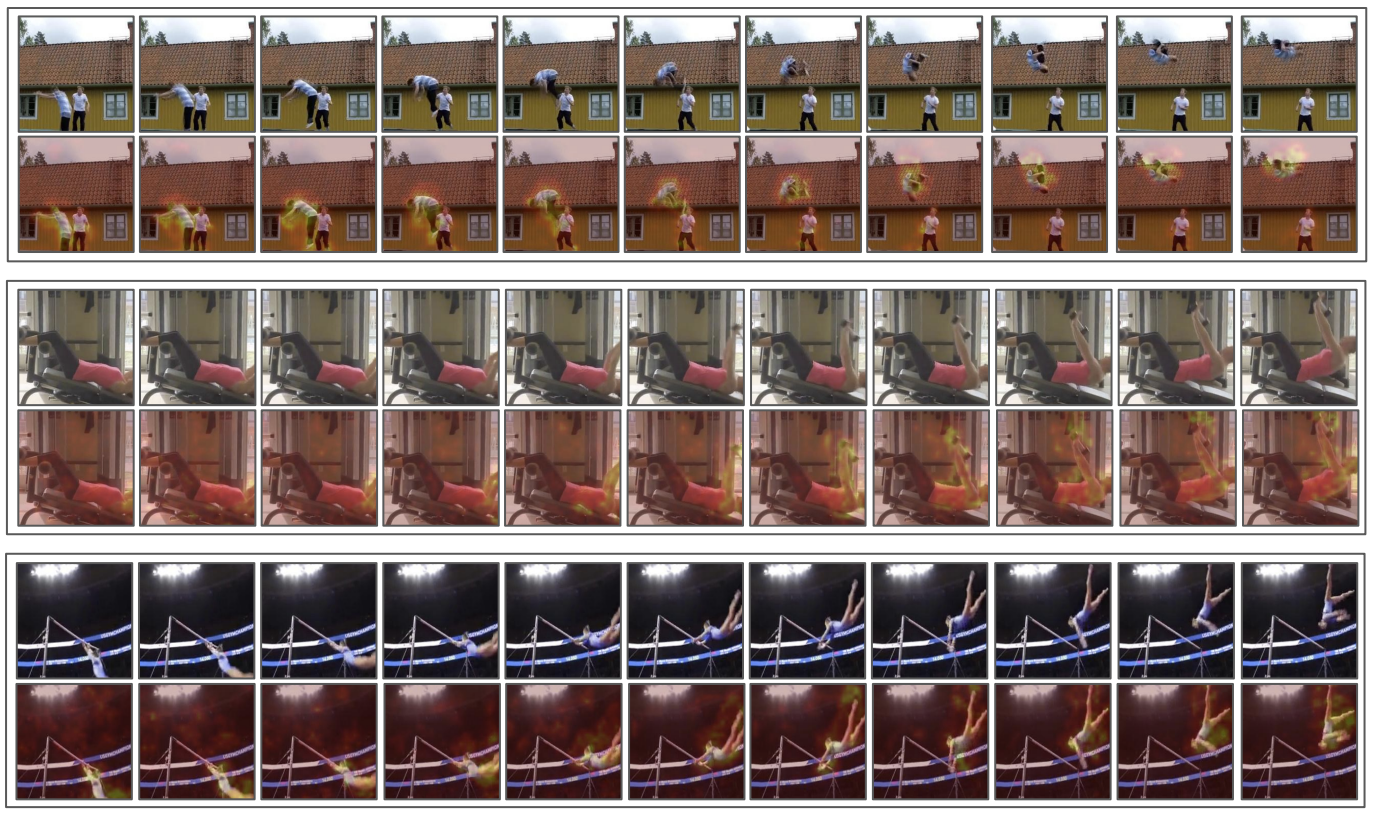}
    \caption{Qualitative Analysis II: visualization of the learned motion confidence map. The \textbf{first two} blocks (categories: jumping on trampoline and situp respectively) are taken from the Kinetics-GEBD dataset, while the \textbf{bottom} block (category: uneven bar) is derived from TAPOS. In each block, the first row shows the RGB frames while the second depicts the motion confidence map learnt by the model. \textbf{Note}: the model is only pre-trained on Kinetics-GEBD but it generalizes to the TAPOS dataset as well.}
    \label{fig:motion}
\end{figure*}

We also perform a qualitative analysis of boundaries detected by our method and compare them with the supervised baseline PC \cite{shou2021generic} and the ground truth annotation in Figure~\ref{fig:bdy}. Figure~\ref{fig:motion} shows a visualization of the motion confidence map learned by the MS module during \textit{pre-training}. We observe that motion confidence generalizes well to the TAPOS dataset too, which was not used for pre-training. This validates that MS module learns general motion features even in a self-supervised setting without any explicit motion specific pretext task. In addition our model's event boundary detection results on the TAPOS dataset further justifies that our model is a generic event boundary detector, as after fine-tuning it generalizes beyond Kinetics-GEBD dataset to the TAPOS benchmark.
We also found that linear evaluation (freezing the encoder) on the downstream GEBD task resulted in poor performance.

\textbf{Note:} For GEBD challenge specific reports, we refer the reader to  challenge page for \href{https://sites.google.com/view/loveucvpr21/program?authuser=0}{2021}, \href{https://sites.google.com/view/loveucvpr22/program?authuser=0}{2022}.

\begin{table}
    \centering
    \caption{Ablation study on validation set of TAPOS and Kinetics-GEBD for F1 score at \textit{Rel.\ Dis} threshold $0.05$}
     \vspace{5pt}
    \begin{tabular}{lll}
    \toprule
        Method & TAPOS & Kinetics-GEBD\\
    \midrule
        Vanilla VCLR & 0.496 & 0.596\\ 
    \midrule
    \rowcolor{Gray}
    cVCLR & 0.502 ($\uparrow$) & 0.605 ($\uparrow$) \\
    \rowcolor{Gray}
        + \textit{MotionSqueeze} & 0.573 ($\uparrow$) & 0.680 ($\mathbf{\uparrow}$)\\
        \rowcolor{Gray}
        + Soft labels  & 0.586 ($\uparrow$)  & 0.711 ($\uparrow$)\\
    \bottomrule
    \end{tabular}
    \vspace{-2mm}
    \label{tab:ablation_study}
\end{table}


\subsection{Ablation Studies}
\noindent \textbf{1: Does \textit{MotionSqueeze} through self-supervision helps?}
As shown in Figure~\ref{fig:motion}, the MS module shows high confidence in regions of  images that are more dynamic. The module learns optical flow in an online fashion even under self-supervision.  Intuitively, temporal order regularization and video segment instance discrimination pre-text tasks implicitly complements the MS module to learn generic motion features. From Table~\ref{tab:ablation_study}, we observe that by incorporating the MS module, the F1@0.05 score on the GEBD task increases by 7.5\% on Kinetics-GEBD and 7.1\% on TAPOS, which is a significant increase. 

\noindent \textbf{2: Does soft labelling helps in boosting the performance?}
Kinetics-GEBD has 5 annotators to capture human perception differences but this introduces ambiguity. Ideally the neighbouring frames of the candidate boundary frame should also have a high value of the ground truth label. To tackle this issue, we use Gaussian smoothing ($\sigma=3$) to create soft labels from hard labels, which ensures that model avoids making over confident predictions for the event boundary. As shown in Table \ref{tab:ablation_study}, soft labels improves the F1@0.05 score by 3.1\% on Kinetics-GEBD and 1.3\% on TAPOS dataset.



\section{Conclusions and Discussion}
In this work, we presented a self-supervised model that can be pre-trained for the generic event boundary detection task.
The GEBD task is an ideal problem for self-supervised learning given that the task aims to learn generic boundaries and is not biased towards any predefined action categories from pre-trained state-of-the-art action recognition models. In order to learn spatial and temporal diversity we reformulate SSL objective at frame-level and clip-level to learn
effective video representations (cVCLR). In addition we augment our encoder with a MS module and find this indeed compliments the overall performance on the downstream GEBD task. Furthermore, the motion features learnt are generic since the model is only pre-trained on Kinetics-GEBD but generalizes to TAPOS dataset as well. Through our extensive evaluation, we achieve comparable performance to self-supervised state-of-the-art methods on the Kinetics-GEBD  as shown in Table~\ref{tab:results_kinetics-gebd}.

However, there are limitations with this work. First, we have not used more powerful models, e.g. transformers as in \cite{li2022structured}, or cascaded networks as in \cite{hong2021generic}. 
Second, since MS module is directly applied on feature maps, it learns global motion features. However, in GEBD the boundaries are generic and every type of motion may not indicate a boundary, hence a more fine-grained motion module can boost the performance. Third, due to computational constraints, our self-supervised model is only pre-trained on the Kinetics-GEBD dataset; however, pre-training the model on Kinetics-400 could yield even better performance on the downstream GEBD task. These limitations will be addressed in our future work. 


\section{Acknowledgement}
This work has emanated from research supported by Science Foundation Ireland (SFI) under Grant Number SFI/12/RC/2289\_P2, co-funded by the European Regional Development Fund and Xperi FotoNation.

{\small
\bibliographystyle{ieee_fullname}
\bibliography{sample}
}

\section{Supplementary : Additional Details}

\subsection{Implementation Details}

\textbf{Stage 2: Finetuning.} 
Input to the encoder is based on the temporal window $W=5$ and  stride $m=3$. $(W, m)$ can be thought of as hyper-parameter, setting a larger value of each might introduce noise information when two different boundaries lie close to each other, a smaller value might be unable to capture the necessary context information for a boundary. Among the 5 annotations available for Kinetics-GEBD for every video, the ones with highest annotator F1 consistency score is used for fine-tuning. 
\subsection{Architectural Design Choice}

Temporal Shift Module (TSM) \cite{lin2019tsm} is inserted in every residual block of ResNet50 encoder. \textit{MotionSqueeze} module is added after the $\texttt{conv3\_x}$ layer of the ResNet50 encoder.

\begin{table}[h]
    
    \centering
    \caption{Modified ResNet50 Encoder}
    \vspace{5pt}
    \label{tab:arch_label}
    \resizebox{8.5cm}{!}{
    \begin{tabular}{c|c|c|c}
    \toprule
    Layers & ResNet-50 & Modified ResNet-50 & Output size \\
    \midrule
    conv1 &  \multicolumn{2}{c|}{$7 \times 7 , 64\text{, stride } 2$} & $112 \times 112$ \\
     \midrule
    & \multicolumn{2}{c|}{$3 \times 3, \text{ max-pool}, \text{stride } 2$} & \\
    \cmidrule{2-3}
    conv2\_x & $\begin{bmatrix} 1 \times 1, 64 \\ 3 \times 3, 64 \\  1 \times 1, 256 \end{bmatrix} \times 3$ & $\begin{bmatrix} \text{TSM} \\ 1 \times 1, 64 \\ 3 \times 3, 64 \\  1 \times 1, 256 \end{bmatrix} \times 3$ & $56 \times 56$ \\
    \midrule
    conv3\_x & $\begin{bmatrix} 1 \times 1, 128 \\ 3 \times 3, 128 \\ 1 \times 1, 128 \end{bmatrix} \times 4$ & $\begin{bmatrix} \text{TSM} \\ 1 \times 1, 128 \\ 3 \times 3, 128 \\ 1 \times 1, 128 \end{bmatrix} \times 4$ & $28 \times 28$ \\
    \midrule
    
    MS Module & \xmark & \multicolumn{1}{|c|}{\cmark} & $28 \times 28$ \\
    \midrule
    
    conv4\_x & $\begin{bmatrix} 1 \times 1, 256 \\ 3 \times 3, 256 \\ 1 \times 1, 1024 \end{bmatrix} \times 6$ & $\begin{bmatrix} \text{TSM} \\ 1 \times 1, 256 \\ 3 \times 3, 256 \\ 1 \times 1, 1024 \end{bmatrix} \times 6$ & $14 \times 14$ \\
    \midrule
    
    conv5\_x & $\begin{bmatrix} 1 \times 1, 512 \\ 3 \times 3, 512 \\ 1 \times 1, 2048 \end{bmatrix} \times 3$ & $\begin{bmatrix} \text{TSM} \\ 1 \times 1, 512 \\ 3 \times 3, 512 \\ 1 \times 1, 2048 \end{bmatrix} \times 3$ & $7 \times 7$ \\
    
    \bottomrule

    \end{tabular}
    }
\end{table}

\noindent It should be noted that our encoder definition is consistent with the architecture design introduced in ResNet \cite{he2016deep} and is different from the encoder in the work of \textit{MotionSqueeze} in \cite{kwon2020motionsqueeze} as shown in Table \ref{tab:arch_label}.

\subsection{Evaluation Protocol.}
We conduct evaluation on two datasets Kinetics-GEBD \cite{shou2021generic} and TAPOS \cite{Shao_2020_CVPR}. For evaluation, we follow the standard evaluation protocol explained in \cite{shou2021generic}, which uses the F1 score as the measurement metric. \textit{Rel.\ Dis} (Relative Distance) is used to decide whether a detected event boundary is correct (if detection probability $\geq 0.5$) or  otherwise incorrect. More formally,  \textit{Rel.\ Dis} is defined as the error between detected and ground-truth timestamps, divided by the length of the whole video. F1 score calculated at \textit{Rel.\ Dis} threshold 0.05 was used as the evaluation metric for the GEBD challenge\textsuperscript{\ref{loveu}}.  We compare our detection results with all annotations (5 annotations per video for Kinetics-GEBD and 1 annotation for TAPOS) in the same video and select the annotation with the highest F1 score. 

\end{document}